\documentclass[num-refs]{nbdt-article}

\usepackage{siunitx}

\usepackage{appendix}
\usepackage{float}
\usepackage[table]{xcolor}
\definecolor{lightgray}{gray}{0.9}
\usepackage{multirow}
\usepackage{makecell}

\papertype{Original Article}

\title{Efficient Deep Reinforcement Learning with\\Predictive Processing Proximal Policy Optimization}

\author[1\authfn{1}]{Burcu Küçükoğlu MSc}
\author[1\authfn{1}]{Walraaf Borkent MSc}
\author[1,2]{Bodo Rueckauer PhD}
\author[1]{Nasir Ahmad PhD}
\author[1]{Umut Güçlü PhD}
\author[1]{Marcel van Gerven PhD}

\contrib[\authfn{1}]{Equally contributing authors.}

\affil[1]{Department of Artificial Intelligence, Donders Institute for Brain, Cognition and Behaviour, Radboud University, Nijmegen, Gelderland, 6525 GD, The Netherlands}
\affil[2]{Department Smart Grids, Fraunhofer Institute for Solar Energy Systems, Freiburg, Baden-W\"urttemberg, 79110, Germany}

\corraddress{Burcu Küçükoğlu MSc, Department of Artificial Intelligence, Donders Institute for Brain, Cognition and Behaviour, Radboud University, Nijmegen, Gelderland, 6525 GD, The Netherlands}
\corremail{burcu.kucukoglu@donders.ru.nl}

\runningauthor{Küçükoğlu et al.}

\begin{document}

\maketitle

\begin{abstract}
Advances in reinforcement learning (RL) often rely on massive compute resources and remain notoriously sample inefficient. In contrast, the human brain is able to efficiently learn effective control strategies using limited resources. This raises the question whether insights from neuroscience can be used to improve current RL methods. Predictive processing is a popular theoretical framework which maintains that the human brain is actively seeking to minimize surprise. We show that recurrent neural networks which predict their own sensory states can be leveraged to minimise surprise, yielding substantial gains in cumulative reward. Spe-cifically, we present the Predictive Processing Proximal Policy Optimization (P4O) agent; an actor-critic reinforcement learning agent that applies predictive processing to a recurrent variant of the PPO algorithm by integrating a world model in its hidden state. 
Even without hyperparameter tuning, P4O significantly outperforms a baseline recurrent variant of the PPO algorithm on multiple Atari games using a single GPU. 
It also outperforms other 
state-of-the-art agents given the same wall-clock time and exceeds human gamer performance on multiple games including Seaquest, which is a particularly challenging environment in the Atari domain.
Altogether, our work underscores how insights from the field of neuroscience may support the development of more capable and efficient artificial agents.

\end{abstract}

\section{Introduction}
The goal of reinforcement learning (RL) is to learn effective control policies based on scalar reward signals provided by the environment. Temporally sparse and delayed reward signals make training an RL agent notoriously slow and unstable. Over the past decade, however, RL has been successfully applied to increasingly complex tasks. This progress is afforded by the use of deep neural networks in combination with algorithmic advances in RL. Research in RL has also been accelerated by the availability of simulation benchmarks that allow rapid testing and comparison of RL algorithms~\citep{bellemare2013arcade}.

The current state of the art in RL is achieved by distributed multi-GPU approaches such as MuZero~\citep{schrittwieser2020mastering}, which utilizes Monte Carlo tree search (MCTS), Agent57~\citep{badia2020agent57}, which combines a large number of innovative approaches into a single model, and GoExplore~\citep{ecoffet2019go}, which keeps an archive of trajectories to force exploration of promising unknown states. However, the high computational cost of these approaches make them infeasible in many research settings.

On the other hand, methods have also been developed for efficient training on single GPU contexts.
These deploy a myriad of strategies from across the range of modern RL research.
For example, Rainbow \citep{hessel2018rainbow} integrates a number of recent developments from Q-learning into a single model, DreamerV2 \citep{hafner2020mastering} relies on world models combined with `imagined' outcomes through predictions of future states, and IQN \citep{dabney2018implicit} efficiently integrates distributional RL techniques with deep Q-learning.
These approaches show promise, however RL remains a sample inefficient and expensive paradigm.

Motivated by the efficiency with which our own brain is able to solve challenging control problems, we ask if we can use brain-inspired algorithms to improve upon the performance of these complex RL approaches. 
\textit{Predictive coding}, an established theory of sensory information processing in the brain \citep{srinivasan1982predictive,mumford1992computational,friston2005theory,clark2013whatever,ciria2021predictive, orlandi2018, huang2011}, proposes that higher-level brain areas attempt to predict the activation of lower-level brain areas and use this prediction to inhibit incoming activity.
The remaining signal, a prediction error, can be seen as a measure of surprise of the internal model of the world that generated the predictions. This surprise signal can be used to adjust an agent's behavior and update its internal model of the world. A number of studies have contributed to the growing experimental evidence for this theory~\citep{alink2010stimulus,naatanen2001primitive,summerfield2008neural,squires1975two,hupe1998cortical,murray2002shape,rao2016circuits,kok2012less,ekman2017time,de2018expectations,dijkstra2020neural,schwiedrzik2017high}, while others reproduced experimentally observed phenomena in explicit computational models of predictive coding~\citep{rao1999predictive,lee2003hierarchical,friston2005theory,friston2010free, spratling2017}.

We hypothesize that simultaneously minimizing sensory surprise and maximizing return (expected cumulative reward) yields more effective and biologically plausible control algorithms.
Specifically, we suppose that by minimizing surprise the agent is forced to learn an internal model which may facilitate learning of more effective control laws.
To test this hypothesis, we leverage recurrent neural network (RNN) models, commonly used to to capture temporal, discrete-time, state evolutions for machine learning and neuroscience ~\citep{jordan1990attractor,elman1990finding,Sussillo2014,maass2016searching,Vyas2020}.
We next investigate whether RNNs that implement predictive processing, hence actively seek to minimize sensory surprise \cite{bubic2010,clark2013whatever}, can learn to efficiently and effectively solve complex RL tasks. To this end, we employ a subset of environments in the Human Atari benchmark, which consists of 57 games where the goal is to beat human-level performance~\citep{bellemare2013arcade}. 

Our results show that game performance of a recurrent variant of the Proximal Policy Optimization (PPO) algorithm~\citep{schulman2017proximal} is strongly improved by including a predictive processing mechanism, yielding a novel Predictive Processing Proximal Policy Optimization (P4O) algorithm. P4O achieves results that are competitive with the current state of the art while using only a fraction of the computational resources.

\section{Methods}

\subsection{P4O architecture}
Our aim is to investigate how predictive processing can aid learning of more effective control policies. For this purpose we augment PPO with a recurrent network and loss function that incorporate a predictive processing error. This approach is motivated by the work of Ali et al.~\citep{ali2021predictive} which demonstrated that error and prediction units as proposed in predictive coding may naturally emerge in energy-constrained neural networks that implement an efficient coding constraint~\citep{barlow1961possible}.

\begin{figure}[!ht]
\centering
\includegraphics[width=0.7\textwidth]{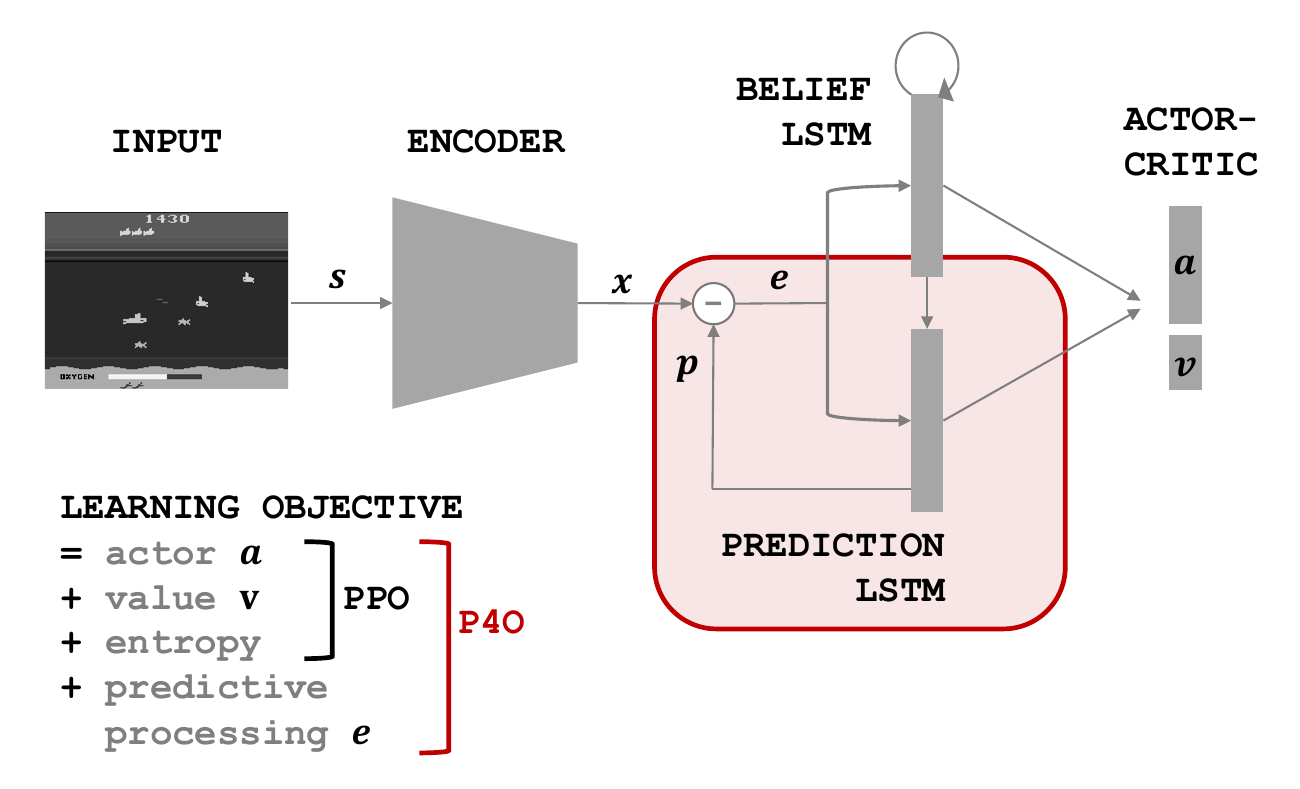}
\caption{Components of the P4O architecture. A sensory input $s$ is encoded into a low-dimensional latent representation $x$. This encoded game state is subtracted from a prediction $p$ generated by an LSTM layer. The resulting prediction error is passed into both the prediction layer and a second LSTM layer representing the agent's belief states. The LSTM outputs are used by an actor-critic model to select an action $a$ and compute a corresponding state value $v$. The red box highlights the main architectural contribution of P4O. Minimization of the prediction error is added to the P4O agent's objective function.}
\label{fig:p4o_architecture}
\end{figure}

A P4O agent consists of three components: an encoder model, a recurrent neural network, and an actor-critic model. The encoder transforms a sensory input into a latent representation. We use a multi-layer CNN with residual connections for this purpose. The recurrent network consists of a modified LSTM layer that incorporates a predictive coding mechanism. The actor-critic component contains one fully connected layer for action selection and one for state value prediction. The agent's objective function is enhanced by a term minimizing the prediction error. The overall architecture is shown in Figure~\ref{fig:p4o_architecture}. The following subsections describe each of these components in more detail.


\subsubsection{Encoder model}
Our encoder is a residual convolutional neural network inspired by Espeholt et al.~\citep{espeholt2018impala}. It consists of four layer groups; each group contains a convolutional layer, a max-pooling layer and two residual blocks with two convolution layers each. The final group is followed by a fully-connected layer with 512 neurons and a $\tanh$ nonlinearity. Layers within a group have the same number of channels; the number of channels across groups increases with the depth of the model (24, 32, 64, and 128 channels, respectively). We use a kernel size of three in all layers, a padding of one and a stride of one. See Appendix~\ref{sec.encodermodelarchitecture} for a visualization of the encoder architecture.

\subsubsection{RNN model}
Our RNN consists of two populations of LSTM cells \citep{hochreiter1997long} which together form the world model of the P4O agent. One population generates a prediction, $p_{t-1}$, of the upcoming latent sensory representation, $x_t$. The other population can be interpreted as a more persistent \textit{belief} state of the agent.
The prediction population's structure is inspired by predictive coding architectures \cite{ali2021predictive,rao1999predictive}, in which a population of error neurons and prediction neurons are separately structured.
The prediction population provides a feedback loop which converts the encoded sensory input into a prediction error signal, $e_t = p_{t-1} - x_t$. 
This prediction error is then used as input to both LSTM cell populations providing the necessary information to update the internal world model.
The prediction LSTM units, acting as information integrators, additionally receive input from the belief LSTM states.

To make the LSTM update rules explicit, we define the states of our belief and prediction LSTM units as $h_t$ and $p_t$, respectively.
These LSTM outputs are controlled by gating variables, $g_{j,t} = \sigma(z_{j,t})$, for the input, output and forget gates, $j \in \{\text{i}, \text{o}, \text{f} \}$.
Here, $z_{j,t} = [z^h_{j,t}, z^p_{j,t}]$ are the internal states of the belief and prediction LSTM unit gates.
In P4O, the gates of the belief population are updated as in regular LSTMs based on error input and the previous hidden state:
\begin{equation}
z_{j,t}^h = W_j^h e_t + U_j^h h_{t-1} + b_j^h \,.
\end{equation}
Here, $W_j$ denotes the input weights, $U_j$ the recurrent weights, and $b_j$ the bias of the corresponding gate. The prediction population likewise receives the prediction error as external input together with the belief states, as discussed above:
\begin{equation}
z_{j,t}^p = W_j^p e_t + U_j^p h_{t-1} + b_j^p \,.
\end{equation}
Notably, the gate variables for the prediction population receive both input from the belief population and indirectly from their own previous state via the prediction error loop.

\subsubsection{Actor-Critic model}
The combined hidden states $[h_t, p_t]$ of the world model are passed as input to an actor-critic model which uses two fully-connected layers to select actions $a_t$ and predict state values $v_t$. The actor layer contains one neuron for each possible action and implements a policy $\pi(a_t \mid h_t)$ by applying a softmax on the network output, resulting in a probability distribution over the action space. 
The agent chooses an action by sampling from this action distribution.
The critic layer consists of a single neuron that outputs the state value $v_t$. The objective function minimized while training the agent is described in the next section.

\subsection{P4O algorithm}

The learning algorithm is based on a modification of the standard PPO algorithm.
This modification makes it suitable for training recurrent models and jointly minimizes prediction error while optimizing for action.
Similar to PPO, at each iteration, the agent retrieves a new batch of data by interacting with a number of parallel environments simultaneously.
The data batch then updates the model by splitting the data into mini-batches and training for multiple epochs while constraining the divergence of the policy.
However, since we use a recurrent model, hidden states are retained during rollout in order to update by backpropagation through time.
A difficulty then arises due to the parameter updates within epochs.
Hidden states become `stale' after any parameter update as they no longer represent the states of the updated RNN.
To avoid this issue, we generate new hidden states by re-running the LSTM after each parameter update.

The loss $L_t$ at time $t$ with respect to the parameters $\theta$ decomposes into a predictive processing loss combined with the other PPO loss terms.
The predictive processing loss is given by 
\begin{equation}
L_{t}^{P}(\theta) =  \sum_{i=1}^{H} \text{MSE}(e_{t+i}) \,,
\label{eq:predloss}\end{equation}
where $e_t$ denotes the prediction error at time $t$, $\text{MSE}(\cdot)$ is a mean squared error, and $H$ is the length of prediction trajectory, allowing prediction error minimization at multiple consecutive timepoints.
If we used a prediction error based only on predicting a single step ahead ($H=1$), the model might be tempted to copy the previous state, since the difference in the environment after a single step can be very small.
To force the world model to learn temporal relationships, we let the model unroll multiple steps ahead ($H>1$) during training and calculate a sequence of predictions while assuming that the input ($x_t$) is precisely cancelled by the prediction produced. This enables an `unaffected' longer time prediction sequence.
We can then compare these predictions to the true RNN inputs for this sequence, measure the prediction errors, and use these errors to compute the loss as per Equation \ref{eq:predloss}.

We combine the predictive processing loss, $L_{t}^{P}(\theta)$,  with standard PPO loss components~\citep{schulman2017proximal} which include an actor loss, $L_{t}^{A}(\theta)$, a critic loss, $L_{t}^{V}(\theta)$, and a loss which penalises low entropy policies (facilitating exploration), $L_{t}^{H}(\theta)$. We sum these to form the combined objective to minimise, defined as
\begin{equation}
L_{t}(\theta) = \hat{ E }_{t}\left[
c_{1} L_{t}^{A}(\theta)+c_{2} L_{t}^{V }(\theta)+c_{3} L_{t}^{P }(\theta)+c_{4} L_{t}^{H}(\theta)
\right] \,,
\label{eq:fulloss}
\end{equation}
where the loss coefficients are denoted by $c_i$ and $\hat{E}_t$ is the empirical mean over a set of samples. 
The actor loss is clipped to avoid strong divergences, as in the original PPO implementation~\citep{schulman2017proximal}.
The critic loss is modified by clipping of the change in value estimate \cite{openaibaselines} and replacement of the squared difference with an absolute difference.
We found these changes to provide an empirical benefit in stability during training.

Unlike standard PPO, we refresh the calculated advantages with the latest model before each update as suggested by~\citep{andrychowicz2020matters}, to prevent basing calculations on old data as in the case with hidden states. The advantages are calculated with generalized advantage estimation~\citep{schulman2015high} in its truncated form, as described by~\citep{schulman2017proximal}. 
Lastly, in standard PPO the first update is unconstrained because the batch was retrieved with the same policy that is being updated, leading always to an action probability ratio of 1.
To prevent the first update from changing the policy too drastically, we ignore the latest policy and instead update based upon the second-to-last policy of the previous batch.

\section{Empirical evaluation}
Here we conduct multiple empirical studies to test our hypotheses around P4O.
\subsection{Hypotheses}
First, we tested whether predictive processing improves the learning of control policies. This is investigated in Section~\ref{sec.performancecomparisons} via performance comparison to a baseline algorithm without predictive processing and a non-surprise minimizing version of P4O within a fixed number of environment steps (Sections~\ref{sec.baselinecomparison} and \ref{sec.individualruns}). We also compare against state-of-the-art single-GPU RL agents within a restricted training time (Section~\ref{sec.sotacomparison}) to measure efficiency of learning both in terms of simulation time steps and wall-clock time.

Second, we tested whether the learning performance of P4O stems from its specific components related to predictive processing or other architectural choices within the LSTM layer. This is investigated via an ablation study (Section~\ref{sec.ablation}), specifically demonstrating the contribution of the following components:
\label{ablationhypothesis} 
\begin{enumerate}
\item[(a)] \textbf{The predictive processing mechanism (the use of prediction error $e_t$ as input into RNN model)}: The removal of this component is achieved by a variant of P4O that uses the direct output of the encoder $x_t$ as input rather than $e_t$, while still keeping prediction as an auxiliary task for maintaining improved internal representations shaped beyond the reward \cite{botvinick2020}. This is implemented through addition of a prediction head at the RNN output, in order to continue minimization of some form of a prediction loss.
\item[(b)] \textbf{The predictive processing loss}: The removal of this component is demonstrated via the removal of this loss term from the objective function in training of the algorithm. 
\item[(c)] \textbf{Combined effects of the predictive processing mechanism and predictive processing loss}: The removal of both is achieved via the baseline algorithm lacking predictive processing.
\item[(d)] \textbf{The belief LSTM layer}: Investigating the role of this component is achieved via dropping the connection from the belief LSTM to the prediction LSTM, which disrupts the influence of belief LSTM on the prediction. Note that this is the only architectural ablation study we can investigate with respect to LSTM cells since removing the belief LSTM directly would give a considerably smaller model complexity, thus disabling a safe comparison when keeping the encoder intact, and removing the prediction LSTM would be already equivalent to (c).
\end{enumerate}

Third, we tested whether the use of predictive processing within P4O leads to different internal representations being learned for the encoder (Section~\ref{sec.inputencoding}). This is demonstrated by a comparison of activation distributions outputted from the encoder in P4O vs. the baseline algorithm without predictive processing.

Finally, we tested whether the use of prediction errors as input in P4O leads to decorrelation of inputs, thus impact how internal representations are learned, in a way that helps performance (Section~\ref{sec.decorr}). Specifically, we test whether specific use of different predictive processing components and architectural choices of P4O lead to different strengths of decorrelation of inputs. This is demonstrated by comparison of the correlation structures of input data before and after the decorrelation ($x_t$ vs. $e_t$) across algorithms.

By the construction of these structured hypotheses, and their empirical evaluation through the ablation studies and comparison to baseline models, we aim to systematically confirm the contribution of our various model components to the performance of P4O.
Such structured investigations have been suggested as a crucial aspect of RL research \cite{patterson2023empirical}, and we attempt, within the computational power available to us, to meet those standards within the aims of this study.

\subsection{Experimental setup}
We utilize the Atari 2600 benchmark commonly used in model-free reinforcement learning research for ease of comparison with state-of-the-art methods. Due to computational resource limitations, we focus on six environments spanning a wide range of difficulty levels. The selected games, from most to least difficult (based on performance of DQN~\cite{mnih2015human} relative to human performance), include Seaquest, Riverraid, Q*bert, Beamrider, SpaceInvaders and Breakout. 
The original Atari frames of 210 by 160 pixels in RGB color were converted to the commonly used 84 by 84 pixels in grayscale format.
We apply the typical four-frame stacking \cite{machado18, Song2020V-MPO}, and therefore the final input is four channels of $84 \times 84$. We use deterministic environments (no sticky actions) with a fixed frame skipping of 4, and
train with 16 environments in parallel. We do not enforce a time or frame limit per episode, as suggested by~\citep{toromanoff2019deep}. The encoder transforms the input $s_t \in \mathbb{R}^{84 \times 84}$ into a low-dimensional input representation $x_t \in \mathbb{R}^p$ with $p=512$. We use a belief LSTM population with state $h_t \in \mathbb{R}^q$ where $q=512$.
Overall, P4O operates with combined predictive and belief LSTM population states $[h_t, p_t] \in \mathbb{R}^k$ with $k=p+q$. Finally, the actor-critic model selects one out of 4-18 possible actions (depending on the game) and generates one state value per time step.

We compare the performance of the P4O algorithm against a recurrent variant of the original PPO algorithm, which we call the LSTM-PPO baseline algorithm. This baseline algorithm also uses a ResNet encoder and is similar to the P4O algorithm in most ways, including the implementation of PPO-related algorithmic details like PPO loss components' calculation, yet lacks predictive processing (see Appendix~\ref{sec.lstmppo} for a diagram).
We used two variants of LSTM-PPO.
First, a model which uses $k=1024$ hidden states, matching the number of hidden states as in the LSTM layer of the P4O architecture.
Second, a model which uses $k=800$ units such that the total number of parameters is comparable with that of the P4O architecture.
Hence, the former keeps the size of the hidden state equal to that of P4O whereas the latter controls for model complexity.
Notably, the discrepancy between scaling of hidden state dimension and model complexity is due to architectural differences between the LSTM-PPO architecture, which has $k^2 + kp + k$ parameters per gate, versus the P4O architecture, with only $kp + kq + k$ parameters per gate.

For most PPO-related hyperparameters we do not apply grid search, but instead use commonly reported hyperparameter values from other PPO implementations (see Appendix~\ref{sec.hyperparams}). For predictive processing, a prediction trajectory length of 3 was chosen by evaluating the trade-off between improved agent performance and increased computational cost.

We additionally ran 
the P4O agent for 10 days to compare performance with the current state-of-the-art in model-based and model-free single GPU agents. A comparison against the LSTM-PPO agent with $k = 800$ hidden states was also provided here to demonstrate the effects of long term training with the inclusion of predictive processing. Note that the training scheme was slightly different for the long runs due to differing learning rate decay schedules.
For additional implementation details, please refer to Appendix~\ref{sec.hardware}. 
All code required to reproduce the simulations described above, is available at 
\url{https://github.com/burcukoglu/P4O-PredictiveProcessingPPO.git}

\subsection{Performance Comparisons}
\label{sec.performancecomparisons}

We here investigate if predictive processing improves the learning of control policies by comparing performance of P4O against both 
some
baseline models and other state-of-the-art RL algorithms.

\subsubsection{Baseline comparison}
\label{sec.baselinecomparison}

\begin{figure}[!ht]
\centering
\includegraphics[width=\textwidth]{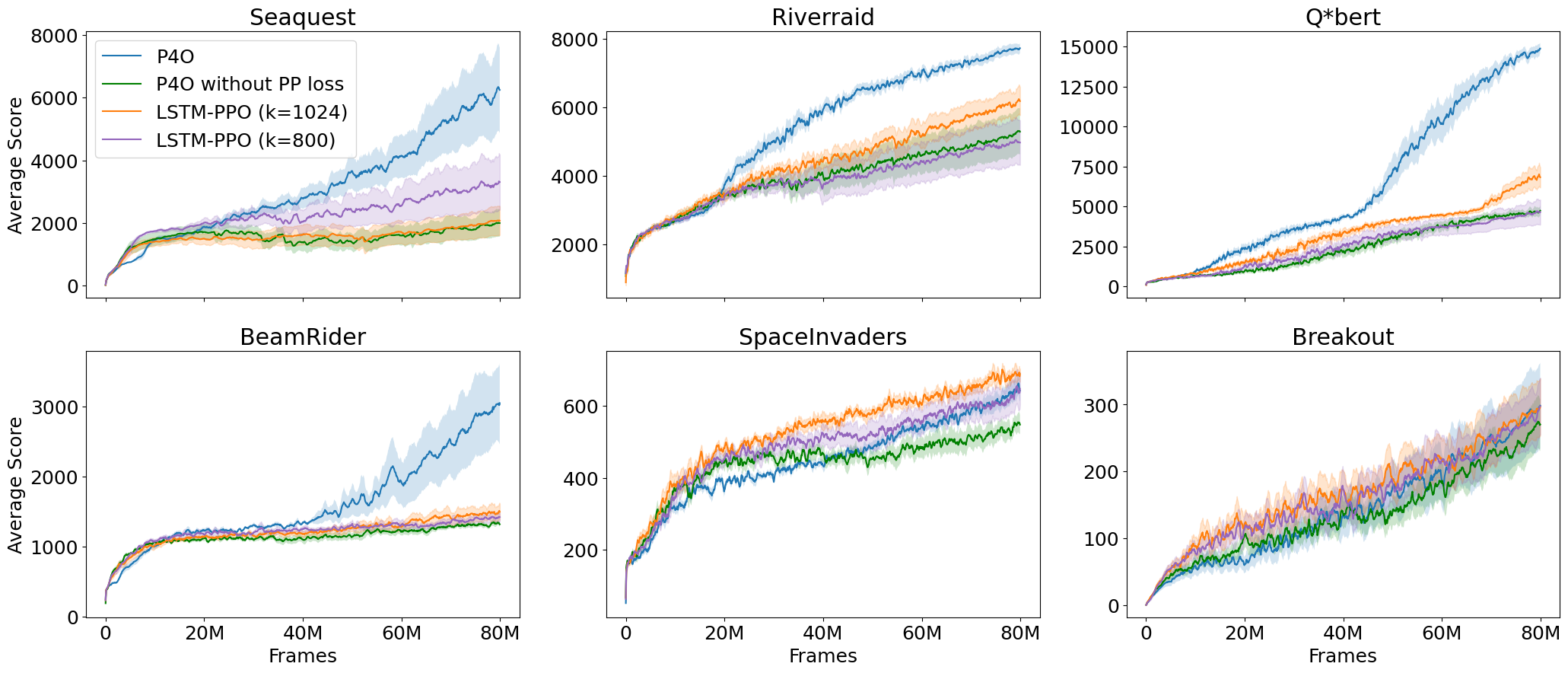}
\caption{Comparison of P4O algorithm against the baselines LSTM-PPO ($k=1024$), LSTM-PPO ($k=800$) and the P4O model optimized without predictive processing loss (P4O without PP loss). Results are based on average score over the last 100 episodes. Shaded areas show standard error of the mean over the 8 runs.}\label{fig:baseresult}
\end{figure}

As Figure \ref{fig:baseresult} demonstrates, the P4O algorithm significantly outperforms the baseline LSTM-PPO algorithm ($k=1024)$ in 4 out of 6 games tested, with on par performance in the remaining games, based on one-tailed t-tests ($p<$ 0.05, $N$ = 16). The difference between the mean learning curves of the two algorithms are statistically significant in all environments except SpaceInvaders and Breakout. The overall effect, except for an insignificant effect for Seaquest, is similar for the LSTM-PPO baseline with $k=800$, which has a comparable number of parameters to the P4O algorithm, confirming that the difference in performance can indeed be attributed to the predictive processing impact, rather than an increase in efficiency due to reduced model complexity. Also, the inclusion of the predictive processing loss in the optimization of the P4O agents yields a significant contribution to performance in all environments ($p<$ 0.05, $N$ = 16) except Breakout, justifying its incorporation into the P4O algorithm. In all environments where the P4O algorithm outperforms the baseline LSTM-PPO algorithm, predictive processing loss was successfully minimized. In SpaceInvaders P4O seems to perform worse, though not significantly, which is associated with a failure to minimize predictive processing loss. In Seaquest, the most difficult game in our chosen set, P4O achieves a mean score of 6238 compared to our LSTM-PPO (k=1024) baseline's mean score of 2070, implying a 3$\times$
increase in performance with the inclusion of predictive processing, while requiring 22\% fewer parameters in total. 
The P4O agent also surpasses the mean score of 1204 reported by the PPO paper~\citep{schulman2017proximal} at 40 million frames in Seaquest.
Note that these results are based on a fixed number of environment steps, where algorithms are yet to converge, with P4O showing even steeper learning progress, hence promise for performance improvement. Further tuning of the hyperparameters for the P4O agent, such as the scaling of the predictive processing component of the loss, may lead to greater performance gains.

\subsubsection{Individual runs}
\label{sec.individualruns}

When inspecting the mean learning curves, we observe a distinct gap between the P4O and baseline agents, especially in Seaquest, Q*bert and BeamRider. This finding indicates some fundamental learning barrier once a certain score level is reached. 
Investigating the behavior of the baseline agents around this saturation point reveals that the agents struggle to integrate the variety of competing goals on different temporal scales. In Seaquest for instance, the agents initially play the game by merely avoiding and destroying the enemy ships while restricting their shooting to the bottom half of the screen. The agents fail to learn to tackle other goals like rescuing the stranded divers or moving up for air when the oxygen level is low. The P4O agents appear to cross this learning barrier, begin to move up for air in time and thus play for much longer, using the entire screen and rescuing divers in the process.
Similarly, games like Q*bert and BeamRider require fulfillment of new goals in each stage before moving on to the next stages, unlike the remaining games. In Q*bert, the agents that learn to jump on each cube twice rather than once at a later stage of the game manage to cross this barrier towards better performance. In BeamRider, high-scoring agents reach Stage 9 by also learning to avoid various enemies, while those only learning to shoot targets get stuck at Stage 2. 

\begin{figure}[!ht]
\centering
\includegraphics[width=\textwidth]{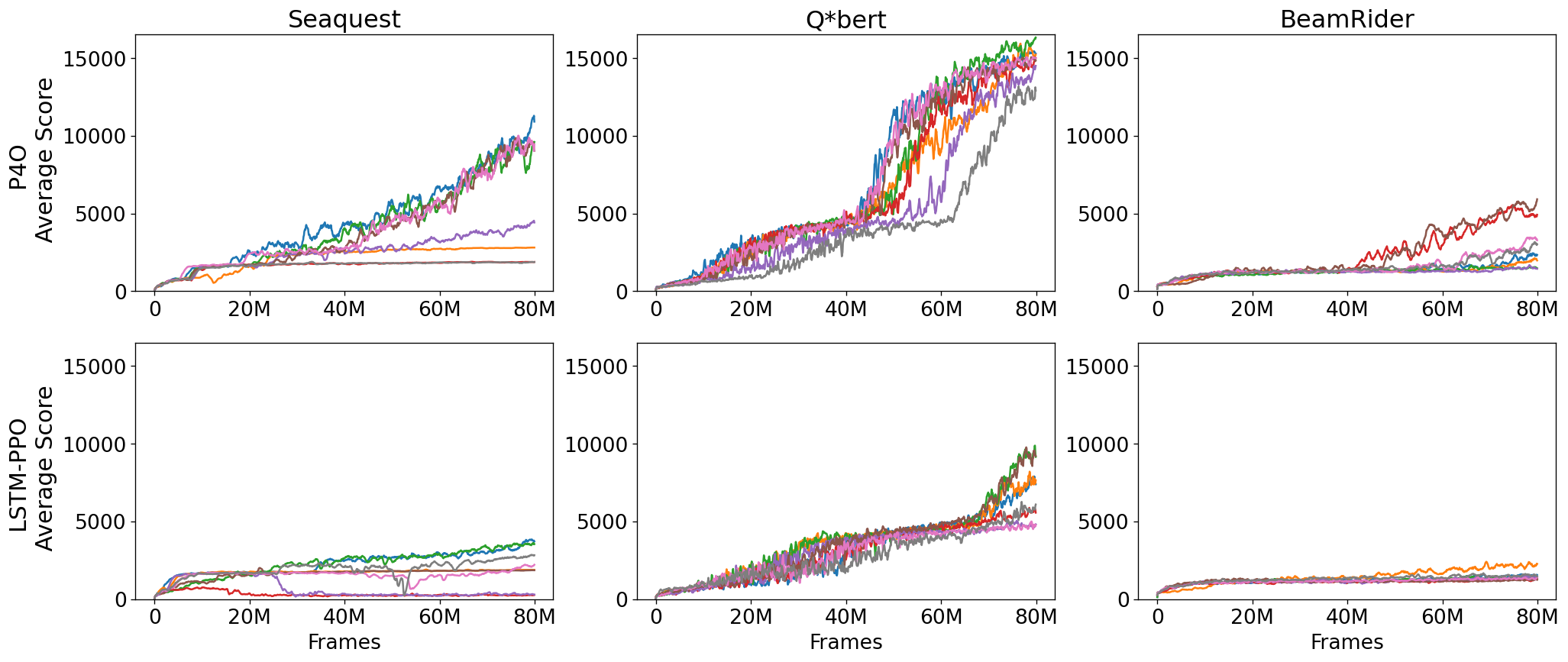}
\caption{Comparison of individual performance curves. P4O results are shown in the top row and LSTM-PPO baseline ($k=1024$) results are shown in the bottom row. Equivalent colors indicate the use of the same seed for network initialization.}\label{fig:seeds}
\end{figure}

Figure~\ref{fig:seeds} shows the score of individual runs for the P4O and LSTM-PPO agents on a variety of games.
Notably, in many cases we can observe that the LSTM-PPO agents reach plateaus in performance due to the above issues around strategy.
Ultimately, this results in a strong negative effect on the final score relative to P4O.

\subsubsection{Comparison with state-of-the-art}
\label{sec.sotacomparison}

\begin{figure}[!ht]
\centering
\includegraphics[width=\textwidth]{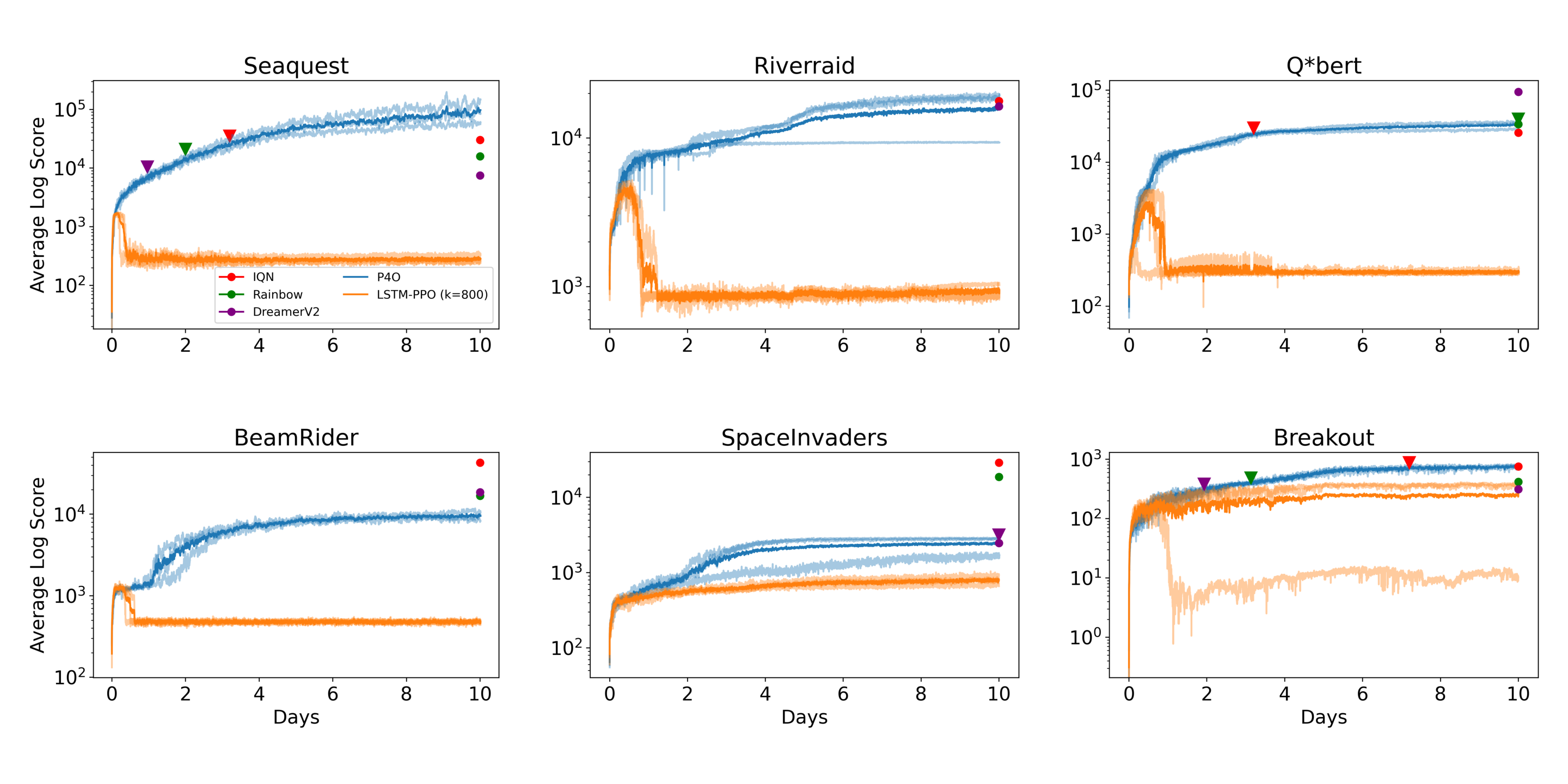}
\caption{P4O's performance comparison against algorithms IQN, Rainbow, DreamerV2 and the baseline LSTM-PPO after 10 days of accelerator time. Lighter curves correspond to the individual runs, whereas darker curves are the mean of the 3 individual runs. Circles represent final reported scores of the IQN, Rainbow and DreamerV2 agents after 10 days. The moment where P4O mean exceeds their respective endpoint score is marked with triangles. Reported average score is in log scale and is based on a rolling mean of the last 100 episodes.} 
\label{fig:compare_long}
\end{figure}

To place the performance of our agent in a broader context, 10 day-long training runs of the P4O agent in all 6 games tested is compared with a number of current state-of-the-art single GPU reinforcement learning agents in Figure \ref{fig:compare_long}. To demonstrate the additional effect of predictive processing, long runs of the baseline LSTM-PPO $(k=800)$ are also provided for comparison against the P4O agent.

The model-based DreamerV2 agent and model-free Rainbow and IQN agents all report their performance after 10 accelerator days in wall-clock time~\citep{hessel2018rainbow,dabney2018implicit,hafner2020mastering}. Each of these models has processed 200M Atari frames at the final test time, whereas our P4O agent is able to process 1.2B frames in 10 days with the same Nvidia Tesla V100 GPU as used by DreamerV2. The DreamerV2 agent uses 22M parameters, whereas the P4O agent uses 7.5M parameters. As can be seen in Figure \ref{fig:compare_long}, P4O performs on par with state-of-the-art algorithms in four of the 17 comparisons, and surpasses them in seven instances. Remarkably, in the game of Seaquest, which is the hardest of them, the P4O agent surpasses the DreamerV2 average after one day, surpasses the Rainbow agent’s final score after roughly two days, and the IQN agent in less than four days, achieving a final average score of 96323 over the last 100 episodes, or 229\% of the human gamer score reported by~\citep{hafner2020mastering}.
P4O also surpasses the human gamer scores in four of the six games (for details see Appendix \ref{sec.scores_table}). 

The performance advantage of the P4O agent may be explained by the inclusion of the predictive processing implementation compared to the LSTM-PPO baseline. When training the P4O agent longer, its performance continues to increases, which is not observed for the LSTM-PPO baseline. In the majority of the games, performance of the LSTM-PPO agent dropped after training roughly longer than a day, as a result of the algorithm diverging from its peak performing policy with further updates. While early stopping could be used for these cases, the peak performance of the P4O agent would still surpass that of LSTM-PPO.

On the other hand, for Seaquest, the performance curve shows no sign of tapering off at the end of even the long run, suggesting that the agent would still benefit from additional time to further approach perfect play. The relatively large gap between the highest score and average score indicates that the agent is still exploring through action sampling with the entropy bonus, although it has already beaten the game multiple times reaching the maximum score (999999). To extract the maximum performance from our agent, we can take the trained agent and run it in a deterministic mode by no longer sampling from the action distribution, but instead always selecting the highest probability action. Testing the trained agent in this way for another 100 episodes leads to a much higher average score of 521684, or 1242\% of the human gamer score.
Further inspection shows that the agent achieved the maximum score in 38\% 
of these episodes. Given this result, it may be beneficial to apply a decay factor to the entropy bonus to allow the model to become more deterministic towards the end of training.

\subsection{Analysis of the P4O algorithm}
We here investigate how P4O's various components contribute to its performance and representation learning.

\subsubsection{Ablation study on algorithm elements}
\label{sec.ablation}
\begin{figure}[!ht]
\centering
\includegraphics[width=\textwidth]{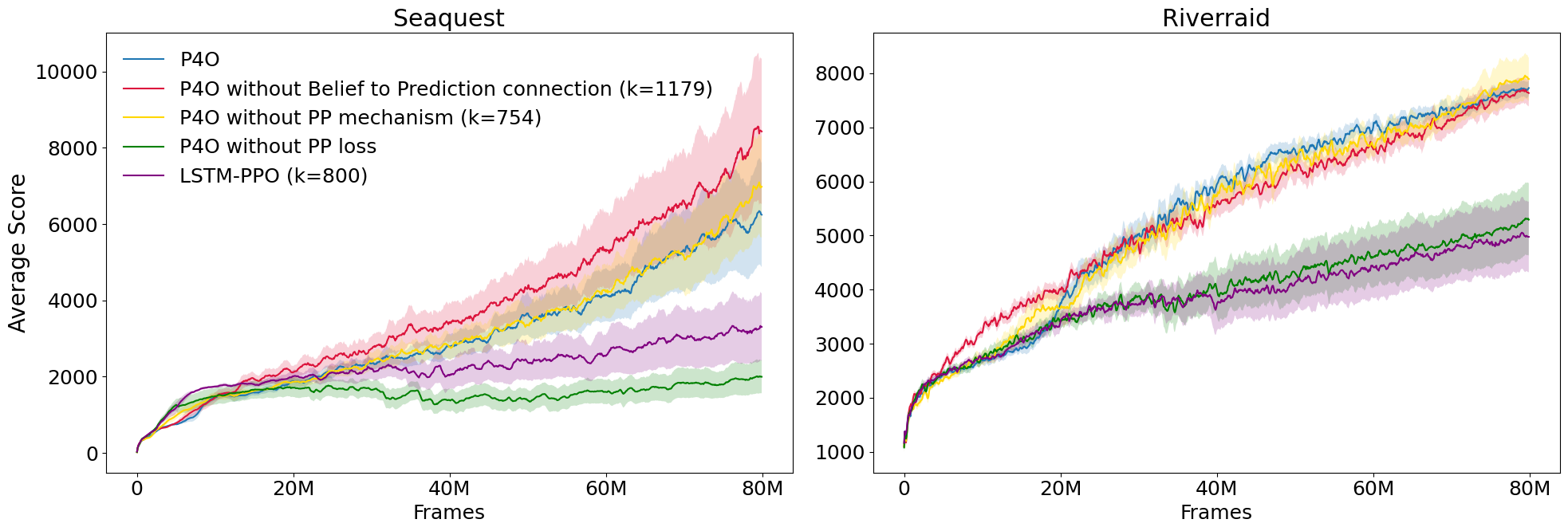}
\caption{Analysis of the contribution of various P4O algorithm elements on performance via ablation study. “P4O without PP Loss” indicates an architecture identical to P4O but which does not include the predictive processing loss in the loss computation. “P4O without PP mechanism” indicates an architecture in which the LSTM hidden states are linearly read-out to predict the input ($x$) with an additional prediction output - but without providing this as a negative feedback signal (hence, no predictive processing mechanism within the network). “P4O without Belief to Prediction connections” indicates a network in which the connections from the belief LSTM to the prediction LSTM units have been cut. Results are based on average score over the last 100 episodes. Shaded areas show standard error of the mean  over the 8 runs.}
\label{fig:ablation}
\end{figure}

Figure~\ref{fig:ablation} demonstrates the contributions of P4O's elements to performance in two of the most difficult games (which use the full action space), with analysis of their significance based on one-tailed t-tests. Note that size of the hidden state $k$ is controlled across the experiments for maintaining a comparable number of total parameters. Following from the elements of interest mentioned in the hypotheses (Section~\ref{ablationhypothesis}):
\begin{enumerate}
\item[(a)] P4O without predictive processing mechanism suggests that the use of prediction error as input by itself does not contribute to a significant improvement on P4O's performance, due to comparable performance levels with P4O ($p>$ 0.05, $N$ = 16).
Note that in this case, the prediction error-based input was removed, but a prediction head was added to the output of the network. Thus, it is clear that, in general, a predictive component to the network’s training is clearly a major benefit to performance \cite{jaderberg2017reinforcement, barreto2017}.
\item[(b)] P4O without the inclusion of any predictive processing loss performs significantly worse than the methods which include a predictive element. This demonstrates the importance of including a predictive processing loss within such an RL framework for maximisation of reward. When this regularizing addition is removed from the objective function, performance drops significantly ($p<$ 0.05, $N$ = 16), as also shown in Section~\ref{sec.baselinecomparison}.
\item[(c)] LSTM-PPO shows that, as also seen in Section~\ref{sec.baselinecomparison}, without any of the predictive processing components (neither the predictive processing loss, nor the prediction error input) the performance drops, significantly especially for Riverraid ($p<$ 0.05, $N$ = 16) in the ablation study.
\item[(d)] P4O without belief LSTM to prediction LSTM connection indicates that the influence of belief LSTM on prediction is not major in contributing to P4O's performance, as comparable performance levels are achieved with and without it ($p>$ 0.05, $N$ = 16). As this connection doesn't have an explicit relation to the predictive processing concept, which is on our focus to incorporate in a control architecture, keeping or dropping such connections can rather be seen as a design choice.
\end{enumerate}

Overall, the ablation study suggests that majority of the performance improvement in P4O comes from the predictive processing loss contributing to a modification of the network's internal model. Furthermore, once a form of prediction loss is included, using prediction error as input additionally doesn't provide significant performance improvements. This may lead to questions regarding the value of the predictive processing mechanism that uses prediction error as input in P4O. The next section investigates this further to see what difference this factor, in combination with other P4O components, may lead to in learning input representations. 

\subsubsection{Input encoding}
\label{sec.inputencoding}

\begin{figure}[!ht]
\centering
\includegraphics[width=\textwidth]{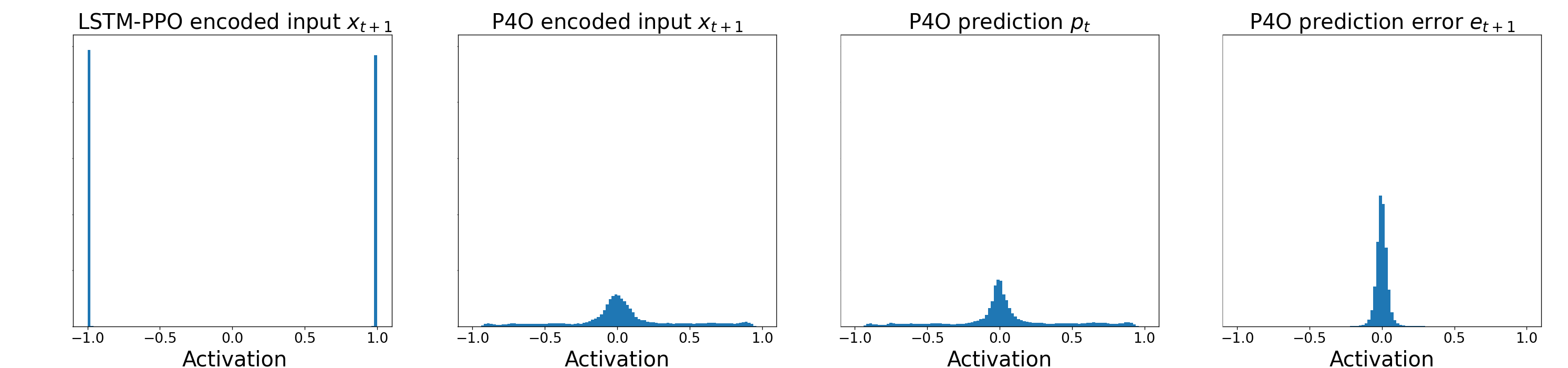}
\caption{Comparison of activation distributions between the LSTM-PPO (k=1024) baseline latent encoding, the P4O latent encoding, the P4O prediction and the P4O prediction error. The distributions are drawn from 1920 states aggregated from a single trained agent on Seaquest.}
\label{fig:act}
\end{figure}

The inclusion of a predictive processing loss term during training has a clear effect on how inputs are encoded (Figure \ref{fig:act}). In the LSTM-PPO baseline algorithm, the output of the encoder follows a typical post-tanh activation profile, with most values being grouped at the extrema (-1 and 1). In contrast, input activations of the P4O algorithm develop a peaked distribution about the origin, despite the final tanh activation layer. Similarly, the prediction $p_{t}$ and prediction error, $e_{t+1}$, distributions are centered around the zero point. The coefficient of determination ($R^2$) of the prediction with respect to the encoded input is 0.987 for the data shown in Figure~\ref{fig:act}, meaning much of the variance in the input is extremely well explained by the prediction of the model.
This demonstrates the effectiveness of incorporating prediction as part of model structure in shaping internal representations of the input.
The activations centered around zero enables economical transmission of the input signal. This is in line with the efficient coding of the predictive coding framework that reduces the dynamic range of the encoded input signal by removing redundancy via economic transmission of only the unpredicted parts \cite{huang2011}.

\subsubsection{Decorrelation of inputs}
\label{sec.decorr}
\begin{figure}[!ht]
\centering
\includegraphics[width=\textwidth]{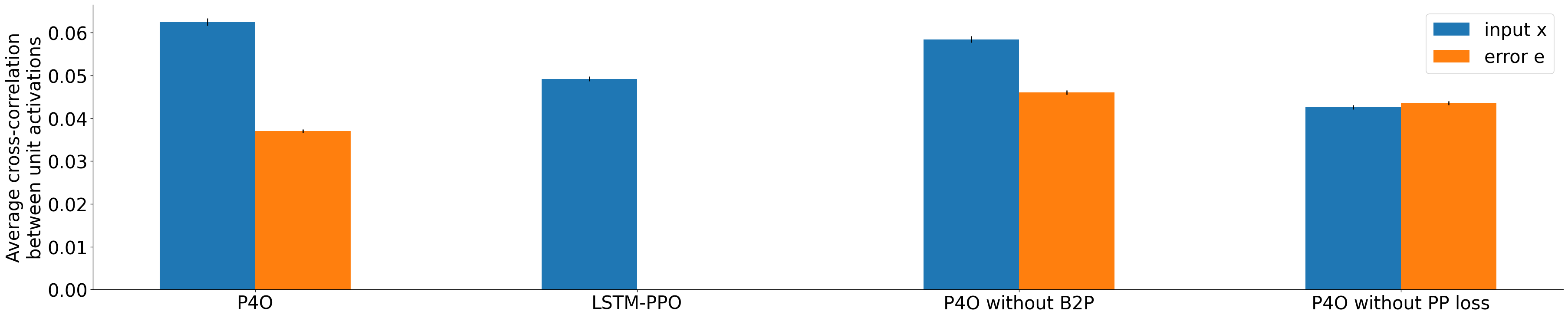}
\caption{Correlation of input data for algorithms investigated in the ablation study. By subtracting the prediction from $x$, the latent representation of the sensory input, P4O uses prediction error $e$ as input to its RNN model, causing decorrelation of the inputs due to error minimization via the predictive processing loss. LSTM-PPO lacks such mechanism, leading to higher input correlation. Plots are based on the upper triangular correlation matrix of inputs for Seaquest runs with a single identical seed. Mean of correlation magnitudes are shown with 95\% confidence intervals.}
\label{fig:decorr}
\end{figure}

Subtracting the prediction away from the input is a form of decorrelation of the inputs \cite{srinivasan1982predictive,huang2011} being provided to the network.
Figure \ref{fig:decorr} shows how the average cross-correlation between activity of the input units to the LSTM layer changes across the P4O variants investigated in the ablation study. Using prediction error as input in P4O indeed leads to a major decrease in correlation in the input.
When using P4O without the predictive processing loss, the agent is not optimized for minimizing this prediction error per-se, hence decorrelation is not in effect.
When using P4O without the connections from belief to prediction LSTMs, despite having the subtractive mechanism for inputs, the prediction LSTM has no access to state information from the belief LSTM to integrate in its prediction process. This results in a lower strength of input decorrelation due to predictions being less informed about dependencies through the belief LSTM. This shows the necessity of keeping these connections for our purposes of incorporating an efficient and non-redundant encoding into control architectures, hence validating our choice for P4O's architecture.
Input decorrelation via the subtractive approach enables a more effective representation of the input internally, and is a requirement considering our incorporation of predictive processing through inspiration from the prediction based inhibition of activity in predictive coding. 

Overall, this study demonstrates the necessity for both the predictive processing mechanism and its optimization via the predictive processing loss with our chosen P4O architecture to be in effect in order to reap the full benefits of predictive processing in learning control policies effectively both for performance and input decorrelation advantages.
Furthermore, methods which include this decorrelated data property are also largely those which perform highly. This suggests that the decorrelation may also aid in the acquiring of higher performance, as has been observed for decorrelated and whitened data in a number of studies \cite{Huang2018-kf, ahmad2023constrained, Luo2017-ku}.

\section{Discussion}

We have here demonstrated that learning of control policies can be significantly improved by incorporating predictive processing within reinforcement learning agents. To this end, we introduced the P4O algorithm, which combines predictive processing with proximal policy optimization in recurrent neural networks. Results show that through the incorporation of a measure of prediction error within the network architecture and loss, the accumulated game score is significantly improved in challenging Atari environments.

Encoding dynamics of agent states in predictive forms has long been a topic of interest to overcome partial observability \cite{littman2001prs,singh2012psr, downey2017psrnn, sutton2015tdn,schlegel2021gvfn}. Rather than predicting observable outcomes \cite{littman2001prs, singh2012psr} or values of multiple observations as in value functions \cite{schlegel2021gvfn,sutton2015tdn}, P4O differs in modelling the dynamics by predicting the next observation's latent representation. Taking these lines of work on constructing compact internal states for prediction further, P4O applies its prediction-based state representation to control settings. This also differentiates P4O from other predictive coding architectures like \cite{lotter2016prednet} that uses predictions for self-supervised learning. P4O is different from other surprise minimising 
or predictive coding  approaches for learning controllable latent dynamics through its simplicity and the use of a recurrent neural network for learning the dynamics model. The simplicity of the method comes from P4O’s use of a subtractive approach as inspired by the inhibition of activity in predictive coding. This approach enables the dynamics model to act not directly on the learned low dimensional latent representation, but rather on its difference from the prediction generated by the dynamics model. Unlike the surprise minimising SMiRL method \cite{berseth2019smirl}, P4O does not need to learn a distribution for latent representations, thus does not require the training of a variational autoencoder. It eliminates the need for decoding from the latent space, which is unnecessary for control. Based on the SMiRL study, one should expect a surprise minimising approach to perform poorly in relatively stable Atari environments with low entropy gap, but P4O does well on the these environments. Learning Controllable Embedding (LCE) algorithms address the problem of applying control in a high-dimensional state space with unknown dynamics by learning a latent dynamics model for direct planning on the embedded lower-dimensional latent space. Within this class of algorithms, approaches like \cite{shu_nguyen_2020_pc3, watter2015e2c, levine2020pcc, solar2018latentreprmodelbasedrl} avoid explicit prediction of high-dimensional observations, just like P4O. This is achieved in~\cite{shu_nguyen_2020_pc3} via a variant of predictive coding called contrastive predictive coding~\cite{vandenoord2018cpc} that focuses on ensuring maximal mutual information between the current and future latent spaces. Application of this ``temporal predictive control"~\cite{nguyen_shu_2021_tpc} along with the Dreamer~\cite{dreamer2020hafneretal} algorithm in the RL setting outperformed other forms of predictive coding such as static predictive coding~\cite{srinivas2020curl} which compares a frame and its state. As can be seen, P4O differs in its definition of predictive coding, taking a more biological approach through a simple subtraction mechanism that does not necessitate comparison of latent states with the future, nor a calculation of mutual information.  Moreover, P4O's implementation of predictive coding decreases the number of parameters needed for the model, saving on compute resources, as opposed to methods that attempt to implement a predictive component by adding extra networks. Compared to other methods of representation learning with additional prediction-based losses~\cite{srinivas2020curl,spr2020dataeffrl_predrepr,pbl2020_bootstraplatentpredictions}, P4O does not require an extension of its architecture, reducing computational cost further.

The current trend in state-of-the-art reinforcement learning has become the use of excessively complex, biologically implausible and computationally intensive multi-GPU agents~\citep{schrittwieser2020mastering, badia2020agent57, ecoffet2019go}. The DreamerV2 agent already made great strides in reversing this trend by reducing complexity and demonstrating what is possible with a single GPU agent~\citep{hafner2020mastering}. Our P4O agent continues this line of research by incorporating a predictive processing element in a standard control architecture.
In a ten-day (wall-clock time) single-GPU training comparison, our P4O agent outperformed other model-free and model-based state-of-the-art reinforcement learning agents.
More impressively, this was accomplished without any hyperparameter optimisation and thus has even greater potential to be uncovered.
Wall-clock time is arguably the most limiting factor in reinforcement learning research, and therefore should be considered as a metric besides cumulative reward. 
Due to the efficient model structure (i.e., three times fewer parameters than DreamerV2), P4O can process frames six times faster than state-of-the-art algorithms DreamerV2~\citep{hafner2020mastering}, IQN~\citep{dabney2018implicit}, and Rainbow~\citep{hessel2018rainbow}. This speedup does not come at the cost of task performance: Within equal training duration, the P4O agent achieves equivalent or higher rewards in 11 out of 17 cases on Atari, including the challenging Seaquest game. One possible limitation here could be the difference in use of sticky actions, which was suggested as a means of introducing stochasticity in training~\cite{machado18}, and used for example by DreamerV2 but not by P4O due to deterministic environmental dynamics. However, the fact that not all algorithms compared provide such experimental details clearly, makes it harder to guarantee similar conditions across comparisons made. Thus still, the competitive performance of P4O demonstrates the power of predictive processing against state-of-the-art algorithms that lack this mechanism.

The question remains as to why RNNs that minimize their prediction error perform so well. We hypothesize that it encourages the RNN to learn an internal representation (world model) of the causes of its sensations~\citep{ha2018recurrent}. Such an induced world model may provide a better basis for control. Furthermore, the use of a predictive processing loss may serve as a regularizer that decorrelates the inputs, which is similar to whitening of sensory input in early visual areas~\cite{Graham2006-qk} and has been shown to have a positive impact on training speed of classification tasks~\cite{Huang2018-kf, ahmad2023constrained, Luo2017-ku}. We expect that a deeper understanding and more widespread adoption of brain-inspired mechanisms such as the one proposed here, will yield more efficient and effective neural controllers in artificial intelligence. 





\section*{Acknowledgements}
This work has received funding from the European Union’s Horizon 2020 research and innovation programme under grant agreement No 899287. This publication is also part of the project Dutch Brain Interface Initiative (DBI2) with project number 024.005.022 of the research programme Gravitation which is (partly) financed by the Dutch Research Council (NWO).



\bibliography{main}



\newpage

\begin{appendices}

\section{Encoder model architecture}
\label{sec.encodermodelarchitecture}

\begin{figure}[H]
\centering
\includegraphics[width=12cm]{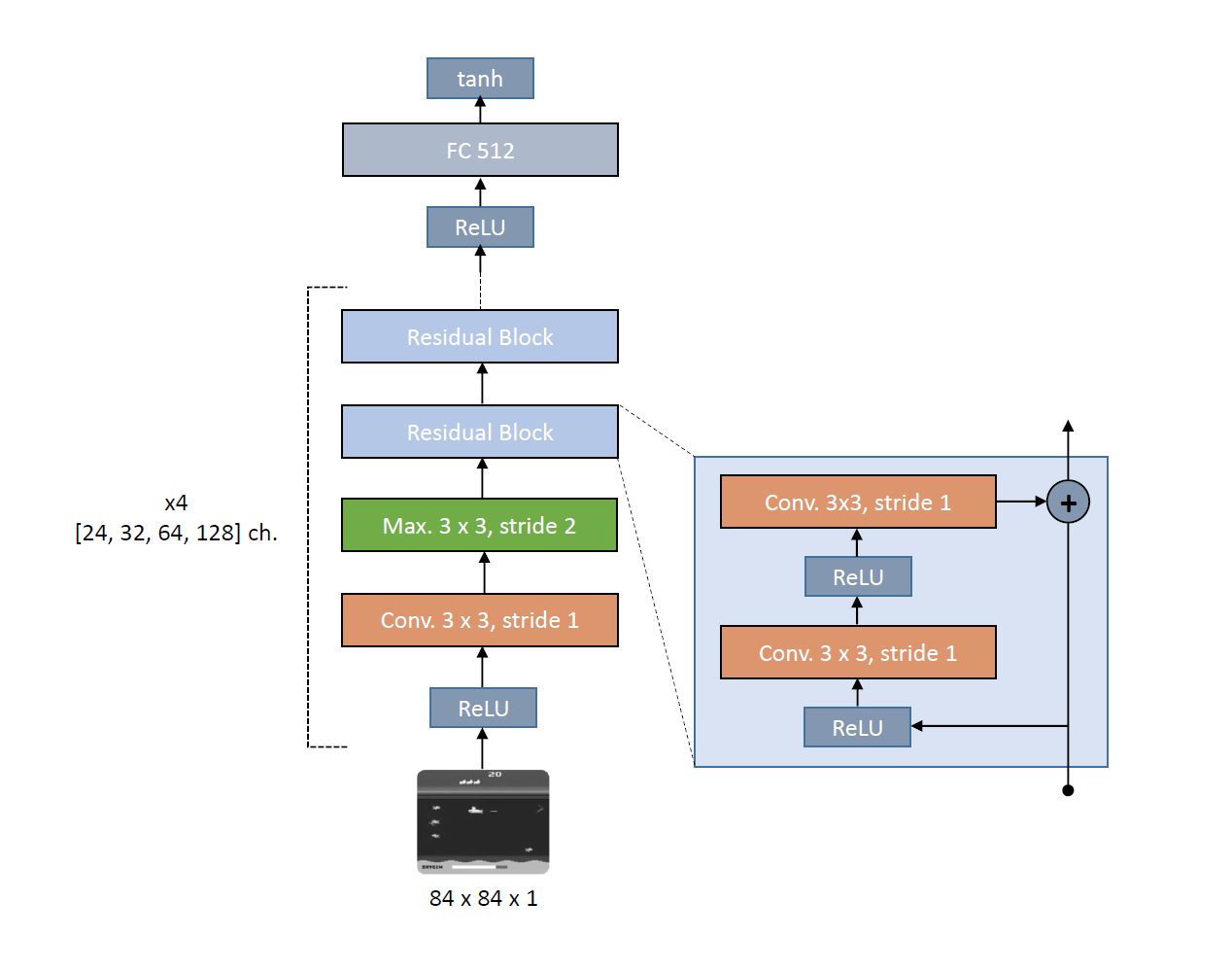}
\caption{Encoder architecture, following a similar structure to the architecture used by~\citep{espeholt2018impala} with a few modifications. The encoder uses a total of 20 convolutional layers and 3.3M parameters.}
\label{fig:enc}
\end{figure}

\section{LSTM-PPO architecture}
\label{sec.lstmppo}

\begin{figure}[H]
\centering
\includegraphics[width=0.7\textwidth]{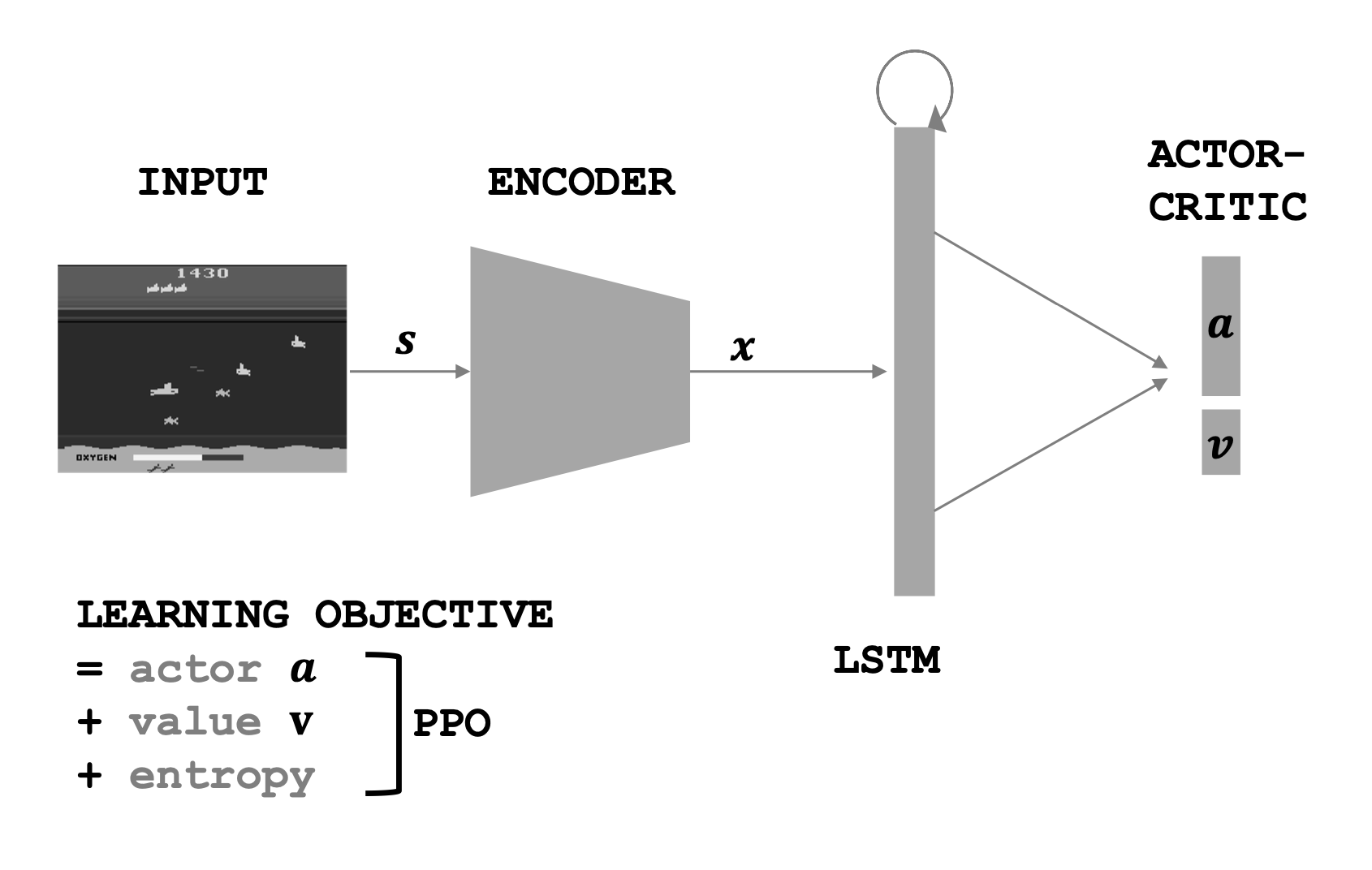}
\caption{Components of the LSTM-PPO architecture. A sensory input $s$ is encoded into a low-dimensional latent representation $x$, which is directly passed into an LSTM. The LSTM output is used by an actor-critic model to select an action $a$ and compute a corresponding state value $v$. Loss minimization is based on the PPO objective. Training scheme of the algorithm is the same with P4O except the details regarding the predictive processing elements which is lacking in LSTM-PPO.}
\label{fig:lstmppo}
\end{figure}

\section{Hyperparameters}\label{sec.hyperparams}
\begin{table}[H]
\caption{Hyperparameters used in the P4O agents. Majority of the PPO-related hyperparameters and algorithmic choices follow after \cite{schulman2017proximal}, e.g. learning rate, optimizer, discount factor, GAE parameter, PPO clip range, or after \cite{openaibaselines}, e.g. epochs per batch, actor loss coefficient, critic loss coefficient.}

\begin{threeparttable}
\small
\begin{tabular}{ll}
\headrow
\tablehead{Hyperparameter} & \tablehead{Value} \\
Learning rate & $2.5 \times 10^{-4}$ \\
Optimizer & Adam \\
Adam $(\epsilon)$ & $1 \times 10^{-5}$ \\
Learning rate decay (long runs) & 0.995 every 100 batches, min. $5 \times 10^{-6}$ \\
Learning rate decay (short runs) & $(1- ($batch number $\times 10^{-4}$)) every batch
, min. $2.5 \times 10^{-8}$ \\
Discount factor $(\gamma)$ & 0.99 \\
GAE parameter $(\lambda)$ & 0.95\\
PPO clip range $(\epsilon)$ & 0.1 \\
Number of parallel environments & 16 \\
Number of batch steps & 125\\
Epochs per batch & 4 \\
Mini-batch size & 400 \\
Number of mini-batches & 5 \\
Actor loss coefficient ($c_{1}$)& 1.0 \\ 
Critic loss coefficient ($c_{2}$)& 0.5 \\ 
Predictive processing loss coefficient ($c_{3}$) & 1.0 \\ 
Entropy term coefficient ($c_{4}$)& 0.02 \\ 
Hidden units in final encoder layer & 512 \\
LSTM hidden units & 1024\\
ResNet channels & [24,32,64,128] \\
Image width, height, channels & 84, 84, 1 \\
Frame stacking & 4 \\
BPTT truncation length & 25 \\
\hline  
\end{tabular}
\end{threeparttable}
\end{table}

\newpage 

\section{Hardware and implementation details} \label{sec.hardware}

We programmed our implementation in Python using the MxNet framework. Because our model is relatively small and efficient, it can be run on a single GPU requiring roughly 6GB of GPU memory. We used a combination of Google Cloud instances with Nvidia Tesla V100 and T4 GPUs and consumer hardware ranging from Nvidia GTX 1060 to RTX 2080TI graphics cards with typical multi-core CPUs to run our experiments. The fact that the agent can be run on an Nvidia GTX 1060 with 6GB of GPU memory demonstrates the small footprint of our model. The choice of CPU did not seem to affect the speed of the model significantly, considering that the largest bottleneck during training was GPU speed. 


\newpage 

\section{Scores for comparison with state of the art } 
\label{sec.scores_table}

\begin{table}[!ht]
\caption{Comparison of our P4O agent with top single GPU agents on all 6 games tested ~\citep{hessel2018rainbow,dabney2018implicit,hafner2020mastering}. Gamer-normalized score based on the human gamer score reported by~\citep{hafner2020mastering}. Scores in parentheses for P4O are achieved when running the trained agent in deterministic mode (only exploitation). P4O's performance is reported in two cases: when trained with the same number of Atari frames and when trained for the same number of accelerator days as the competitors.}
\begin{threeparttable}
\small
\begin{tabular}{lccccc}
\headrow
\tablehead{Game} & \tablehead{Agent} & \tablehead{Atari Frames} & \tablehead{Accelerator Days} & \tablehead{Average Score} & \tablehead{Gamer-Normalized Score}\\
\hiderowcolors
\multirow{5}{*}{\small{Seaquest}} & \small{DreamerV2} & \small{200M} & \small{10} & \small{7480} & \small{0.18} \\ 
& \small{Rainbow} & \small{200M} & \small{10} & \small{15898} & \small{0.38} \\ 
& \small{IQN} & \small{200M} & \small{10} & \small{30140} & \small{0.72} \\
 & \small{P4O} & \small{200M} & \small{1.7} & \small{10815} & \small{0.26} \\ 
& \textbf{\small{P4O}} & \textbf{\small{1.2B}} & \textbf{\small{10}} & \textbf{\small{96323 (521684)}} & \textbf{\small{2.29 (12.42)}} \\

\Xhline{0.1\arrayrulewidth}
\multirow{5}{*}{\small{Riverraid}} & \small{DreamerV2} & \small{200M} & \small{10} & \small{16351} & \small{0.96} \\ 
& \small{Rainbow} & \small{200M} & \small{10} & \small{-} & \small{-} \\ 
& \small{IQN} & \small{200M} & \small{10} & \small{17765} & \small{1.04} \\
 & \small{P4O} & \small{200M} & \small{1.7} & \small{8056} & \small{0.47} \\
& \textbf{\small{P4O}} & \textbf{\small{1.2B}} & \textbf{\small{10}} & \textbf{\small{15639 (23820)}} & \textbf{\small{0.91 (1.39)}} \\

\Xhline{0.1\arrayrulewidth}
\multirow{5}{*}{\small{Q*bert}} & \small{DreamerV2} & \small{200M} & \small{10} & \small{94688} & \small{7.04} \\ 
& \small{Rainbow} & \small{200M} & \small{10} & \small{33817} & \small{2.51} \\ 
& \small{IQN} & \small{200M} & \small{10} & \small{25750} & \small{1.91} \\
 & \small{P4O} & \small{200M} & \small{1.7} & \small{15567} & \small{1.16} \\
& \textbf{\small{P4O}} & \textbf{\small{1.2B}} & \textbf{\small{10}} & \textbf{\small{33054 (31075)}} & \textbf{\small{2.46 (2.31)}} \\

\Xhline{0.1\arrayrulewidth}
\multirow{5}{*}{\small{BeamRider}} & \small{DreamerV2} & \small{200M} & \small{10} & \small{18646} & \small{1.10} \\ 
& \small{Rainbow} & \small{200M} & \small{10} & \small{16850} & \small{1.00} \\ 
& \small{IQN} & \small{200M} & \small{10} & \small{42776} & \small{2.53} \\
 & \small{P4O} & \small{200M} & \small{1.7} & \small{3007} & \small{0.18} \\
& \textbf{\small{P4O}} & \textbf{\small{1.2B}} & \textbf{\small{10}} & \textbf{\small{9566 (11439)}} & \textbf{\small{0.57 (0.68)}} \\

\Xhline{0.1\arrayrulewidth}
\multirow{5}{*}{\small{SpaceInvaders}} & \small{DreamerV2} & \small{200M} & \small{10} & \small{2474} & \small{1.48} \\ 
& \small{Rainbow} & \small{200M} & \small{10} & \small{18789} & \small{11.26} \\ 
& \small{IQN} & \small{200M} & \small{10} & \small{28888} & \small{17.31} \\
 & \small{P4O} & \small{200M} & \small{1.7} & \small{759} & \small{0.46} \\
& \textbf{\small{P4O}} & \textbf{\small{1.2B}} & \textbf{\small{10}} & \textbf{\small{2515 (3035)}} & \textbf{\small{1.51 (1.82)}} \\

\Xhline{0.1\arrayrulewidth}
\multirow{5}{*}{\small{Breakout}} & \small{DreamerV2} & \small{200M} & \small{10} & \small{312} & \small{10.4} \\ 
& \small{Rainbow} & \small{200M} & \small{10} & \small{417} & \small{13.9} \\ 
& \small{IQN} & \small{200M} & \small{10} & \small{754} & \small{25.13} \\
 & \small{P4O} & \small{200M} & \small{1.7} & \small{279} & \small{9.29} \\
& \textbf{\small{P4O}} & \textbf{\small{1.2B}} & \textbf{\small{10}} & \textbf{\small{758 (864)}} & \textbf{\small{25.27 (28.80)}} \\
\hline  
\end{tabular}

\end{threeparttable}
\end{table}

\end{appendices}

\end{document}